\documentclass{article}

\PassOptionsToPackage{numbers, compress}{natbib}

\usepackage[preprint]{neurips_2026}

\usepackage[utf8]{inputenc} 
\usepackage[T1]{fontenc}    
\usepackage{hyperref}       
\usepackage{url}            
\usepackage{booktabs}       
\usepackage{amsfonts}       
\usepackage{nicefrac}       
\usepackage{microtype}      
\usepackage[table]{xcolor}         

\usepackage{amsmath}
\usepackage{amsthm}
\usepackage{amssymb}
\usepackage{booktabs}
\usepackage{bm}
\usepackage[inline]{enumitem}
\usepackage{listings}
\usepackage{csquotes}
\usepackage{nicefrac}
\usepackage{pgfplots}
\usepackage{subcaption}
\usepackage{todonotes}
\usepackage{comment}
\usepackage{wrapfig}

\usepackage{tikz}
\usepackage{graphicx}
\usetikzlibrary{arrows.meta, positioning, calc}
\pgfplotsset{compat=1.18}

\makeatletter
\renewcommand{\paragraph}{%
  \@startsection{paragraph}{4}{\z@}%
                {0.5ex \@plus 0.5ex \@minus 0.2ex}%
                {-1em}%
                {\normalsize\bf}%
}
\makeatother

\newtheorem{definition}{Definition}
\newtheorem{proposition}{Proposition}

\renewcommand{\cite}[1]{\citep[][]{#1}}
\newcommand{\egcite}[1]{\citep[e.g.,][]{#1}}
\newcommand{\inlinecite}[1]{\citet{#1}}

\newcommand{\suc}{\ensuremath{\textit{succ}}}

\newcommand{\pspace}{\textbf{\textup{PSPACE}}}

\newcommand{\expspace}{\textbf{\textup{EXPSPACE}}}

\newcommand{\denselist}{\itemsep 0pt\partopsep 0pt}

\newcommand{\agent}[1]{\todo[inline]{Agent: #1}}
\newcommand{\augusto}[1]{\todo[inline,color=blue!35]{Augusto: #1}}
\newcommand{\jendrik}[1]{\textcolor{red}{Jendrik: #1}}
\newcommand{\andre}[1]{\todo[color=green!35]{Andr\'{e}: #1}}
\renewcommand{\agent}[1]{}
\renewcommand{\jendrik}[1]{}
\renewcommand{\augusto}[1]{}
\renewcommand{\andre}[1]{}

\definecolor{mycolor0}{rgb}{0.12156862745098039, 0.4666666666666667, 0.7058823529411765}
\definecolor{mycolor1}{rgb}{1.0, 0.4980392156862745, 0.054901960784313725}
\definecolor{mycolor2}{rgb}{0.17254901960784313, 0.6274509803921569, 0.17254901960784313}
\definecolor{mycolor3}{rgb}{0.8392156862745098, 0.15294117647058825, 0.1568627450980392}
\definecolor{mycolor4}{rgb}{0.5803921568627451, 0.403921568627451, 0.7411764705882353}
\definecolor{mycolor5}{rgb}{0.5490196078431373, 0.33725490196078434, 0.29411764705882354}
\definecolor{mycolor6}{rgb}{0.8901960784313725, 0.4666666666666667, 0.7607843137254902}
\definecolor{mycolor7}{rgb}{0.4980392156862745, 0.4980392156862745, 0.4980392156862745}
\definecolor{mycolor8}{rgb}{0.7372549019607844, 0.7411764705882353, 0.13333333333333333}
\definecolor{mycolor9}{rgb}{0.09019607843137255, 0.7450980392156863, 0.8117647058823529}

\title{Property-Guided LLM Program Synthesis\\for Planning}

\author{
Andr\'{e} G.\ Pereira\\
Federal University of Rio Grande do Sul\\
Brazil
\And
Augusto B.\ Corr\^{e}a\\
University of Oxford\\
United Kingdom
\And
Jendrik Seipp\\
Link{\"o}ping University\\
Sweden
}

\begin{document}

\maketitle

\begin{abstract}
LLMs have shown impressive success in program synthesis, discovering programs that surpass previously known solutions. However, these approaches rely on simple numeric scores to signal program quality, such as the value of the solution or the number of passed tests. Because such a score offers no guidance on \emph{why} a program failed, the system must generate and evaluate many candidates per problem in the hope that some succeed, increasing both LLM inference and evaluation costs.
We study a different approach: property-guided LLM program synthesis. Instead of scoring programs after evaluation, we check whether a candidate satisfies a formally defined property. When the property is violated, we stop the evaluation early and provide the LLM with a concrete counterexample showing exactly how the program failed. This feedback drastically reduces both the number of program generations and the evaluation cost, and can guide the LLM to generate stronger programs. We evaluate this approach on PDDL planning domains, asking the LLM to synthesize \emph{direct} heuristic functions: every state reachable by strictly improving transitions has a strictly improving successor. A heuristic with this property leads hill-climbing algorithm directly to a goal state. A counterexample-guided repair loop generates one candidate program, checks the property over a training set, and returns the first case that violates the property. We evaluate our approach on ten planning domains with an out-of-distribution test set. The synthesized heuristics are effectively \emph{direct} on virtually all test tasks, and compared to the best prior generation method our approach generates seven times fewer programs per domain on average, solves more tasks without using search, and requires several orders of magnitude less computation to evaluate candidates. Whenever a problem admits a verifiable property, property-guided LLM program synthesis can reduce synthesis and evaluation cost while improving program quality.
\end{abstract}

\section{Introduction}

Using large language models (LLMs) for \emph{program synthesis} yields programs that surpass previously known solutions on problems ranging from competitive programming~\cite{li-et-al-arxiv2022} to open mathematical questions~\cite{romera-paredes-et-al-nature2024} and algorithmic discovery~\cite{novikov2025-arXiv2025,liu-icml2024,tuisov-et-al-icaps2026,correa-et-al-neurips2025,chen-arXiv2025}.
These approaches use simple numeric scores to rank candidate programs, such as the average value of the solution or the number of passed tests. Because such a score offers no guidance on \emph{why} a program failed, the system must generate and evaluate many candidates per problem in the hope that some succeed, increasing both LLM inference cost and evaluation cost.

There is an idea from formal methods that replaces this fitness score with something more informative: \emph{counterexample-guided inductive synthesis} (CEGIS)~\cite{solar-lezama-pldi2005,jha-icse2010,abate-cav2018}. A verifier checks each candidate against a formal specification, and on failure it returns a concrete input on which the candidate is wrong; the synthesizer can then repair or discard the candidate using that witness instead of a single number. CEGIS underpins much of classical program synthesis, yet it has seen little use with LLMs.
We use an LLM as the synthesizer in a CEGIS-style loop with a property checker as the verifier. On every failed candidate, we hand the LLM a concrete counterexample of where and how the program went wrong, which lets us stop evaluation as soon as the property is violated. This reduces the number of candidates the LLM has to generate and steers it toward stronger programs.

We evaluate this idea on \emph{classical planning}~\cite{ghallab-et-al-2004}, a problem known to be challenging for LLMs~\cite{valmeekam-et-al-neurips2023,valmeekam-et-al-neurips2023datasets,stechly-et-al-neurips2024}. Given a planning domain, we ask the LLM to synthesize a \emph{heuristic function} as Python code, which is then automatically injected into a planner. Ideally, such a heuristic would guide hill climbing from the initial state to a goal without ever getting stuck, solving tasks without any combinatorial search. To capture this requirement formally, we target the \emph{direct} property~\cite{dold-helmert-aaai2024}: a heuristic is direct if every state reachable from the initial state via strictly improving transitions has a strictly improving successor. This property is a natural fit for property-guided synthesis, since any violation immediately yields a concrete counterexample. Synthesizing a direct heuristic is challenging both in theory \cite{helmert-et-al-icaps2022,dold-helmert-aaai2024} and in practice \egcite{frances-et-al-ijcai2019}: it requires reasoning about every state on every strictly improving path, and a single state without an improving successor is enough to break the property.
In fact, synthesizing a direct heuristic for a given task is as hard as solving the task itself.

\begin{figure}[t]
\centering

\tikzset{
    fig/.style = {draw, thick, rounded corners, fill=blue!5, inner sep=1ex, align=center, text width=2.2cm, minimum height=1.2cm, font=\scriptsize},
    caption/.style = {font=\scriptsize, align=center},
    ->/.style = {-{Latex[length=2mm,width=1.5mm]}, line width=0.6pt},
}

\begin{tikzpicture}[node distance=1.2cm and 0.8cm]

\node[fig, fill=orange!10,] (input) {
    \textbf{Initial Prompt}\\
    Domain \&\\
    Training Task
};

\node[fig, fill=black!10, right=0.8cm of input] (llm) {
    \textbf{LLM}\\
    Program Synthesis
};

\node[fig,  right=0.8cm of llm] (heuristic) {
    \textbf{Candidate Heuristic}\\
    (Python Code)
};

\node[fig, right=0.8cm of heuristic] (validator) {
    \textbf{Validator}\\
    Checks Direct\\
    Property
};

\node[fig, fill=red!10, below=.8cm of llm] (counterexample) {
    \textbf{Counterexample Prompt}\\
    Failing State and Successors
};

\node[fig, fill=green!10, below=.8cm of validator] (success) {
    \textbf{Verified}\\
    Direct\\
    Heuristic
};

\draw[->] (input.east) -- (llm.west);
\draw[->] (llm.east) -- (heuristic.west);
\draw[->] (heuristic.east) -- (validator.west);

\draw[->] ([xshift=10pt]validator.south) -- node[right, font=\scriptsize] {Pass} ([xshift=10pt]success.north);

\draw[->] (validator.south west) -- node[below, font=\scriptsize, sloped, midway] {Fail (Property Violated)} (counterexample.east);

\draw[->] (counterexample.north) -- node[left, font=\scriptsize, midway] {Repair} (llm.south);

\end{tikzpicture}
\caption{Counterexample-driven repair loop for property-guided heuristic function synthesis. 
Given a planning domain and a task from the training set, the LLM proposes a candidate heuristic function in Python. 
The validator evaluates this heuristic on the training set, verifying the \emph{direct} property. 
If the property is satisfied, the valid heuristic is output. 
If the property is violated, evaluation stops early, and a concrete counterexample is returned.}
\label{fig:method}
\end{figure}
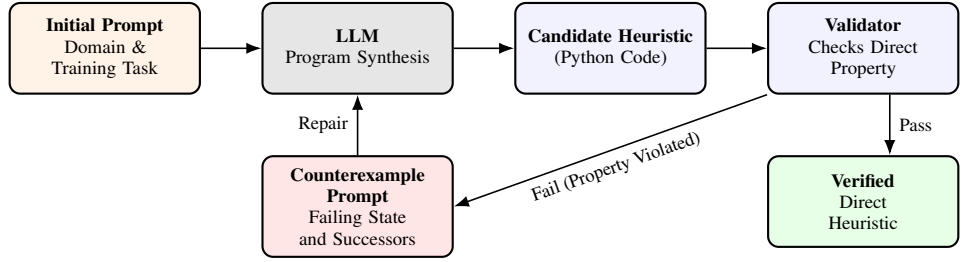

Our counterexample-driven repair loop takes a PDDL (Planning Domain Definition Language) domain together with a training set of PDDL tasks from this domain as input, and validates the direct property over the training set.
Validation terminates as soon as a counterexample is found. In this case, we prompt the LLM with the failing task and the counterexample: the state where the property fails, its heuristic value, and the successor states with their heuristic values. The LLM is asked to provide a new heuristic that is direct for this task. This loop repeats until the heuristic passes the full training set. Although validation is confined to the training tasks, we aim for the synthesized heuristic to generalize, remaining direct on the much harder, out-of-distribution test tasks. Figure~\ref{fig:method} illustrates this process. To our knowledge, our work is the first to use LLMs to synthesize heuristics via property-guided repair.

On the ten planning domains from the Learning Track of the International Planning Competition (IPC) 2023~\cite{taitler-et-al-aimag2024}, evaluated on an out-of-distribution test set, our approach generates seven times fewer programs per domain on average compared to the best prior generation method~\cite{correa-et-al-neurips2025}, requires significantly less computation to evaluate candidates, and solves more tasks via hill climbing without any combinatorial search. Our repair loop always finds a heuristic that is direct on every training task within the iteration budget, and the synthesized heuristics remain direct on virtually all out-of-distribution test tasks, confirming that the property generalizes beyond the training set.

\section{Background}

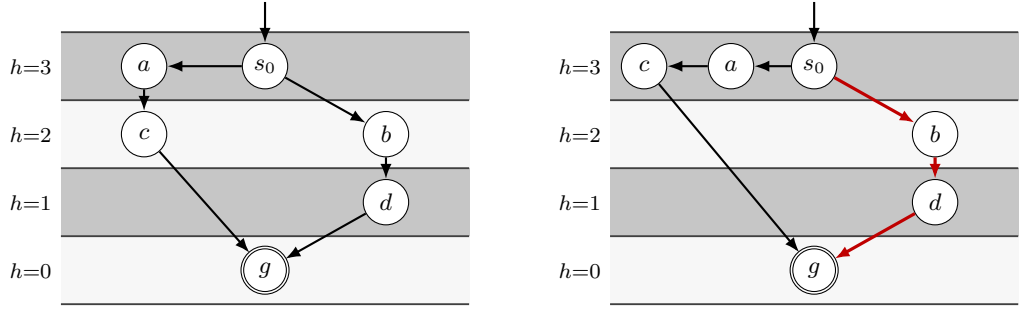
\begin{figure}[t]
\centering
\begin{subfigure}[t]{0.48\linewidth}
\centering
\begin{tikzpicture}[
  every node/.style={font=\small},
  state/.style={circle, draw, minimum size=6mm, inner sep=0pt, fill=white},
  goal/.style={state, double, fill=white},
  arrow/.style={-{Latex[length=2mm]}, thick}
]
\fill[gray!45] (-2.7,2.25) rectangle (2.7,3.15);
\fill[gray!6]  (-2.7,1.35) rectangle (2.7,2.25);
\fill[gray!45] (-2.7,0.45) rectangle (2.7,1.35);
\fill[gray!6]  (-2.7,-0.45) rectangle (2.7,0.45);
\foreach \y in {-0.45,0.45,1.35,2.25,3.15} {\draw[black!75, line width=0.7pt] (-2.7,\y) -- (2.7,\y);}

\node[anchor=east, font=\footnotesize] at (-2.7,2.7) {$h{=}3$};
\node[anchor=east, font=\footnotesize] at (-2.7,1.8) {$h{=}2$};
\node[anchor=east, font=\footnotesize] at (-2.7,0.9) {$h{=}1$};
\node[anchor=east, font=\footnotesize] at (-2.7,0)   {$h{=}0$};

\node[state] (s0) at (0,2.7) {$s_0$};
\node[state] (a)  at (-1.6,2.7) {$a$};
\node[state] (b)  at ( 1.6,1.8) {$b$};
\node[state] (c)  at (-1.6,1.8) {$c$};
\node[state] (d)  at ( 1.6,0.9) {$d$};
\node[goal]  (g)  at ( 0,0) {$g$};

\draw[arrow] (0,3.55) -- (s0);
\draw[arrow] (s0) -- (a);
\draw[arrow] (s0) -- (b);
\draw[arrow] (a)  -- (c);
\draw[arrow] (b)  -- (d);
\draw[arrow] (c)  -- (g);
\draw[arrow] (d)  -- (g);
\end{tikzpicture}
\caption{Descending heuristic: every alive state has at least one strictly improving successor, including state $a$ which will never be expanded by GBFS or HC.}
\label{fig:desc-descending}
\end{subfigure}\hfill
\begin{subfigure}[t]{0.48\linewidth}
\centering
\begin{tikzpicture}[
  every node/.style={font=\small},
  state/.style={circle, draw, minimum size=6mm, inner sep=0pt, fill=white},
  goal/.style={state, double, fill=white},
  plateau/.style={state, draw=red!75!black, very thick},
  arrow/.style={-{Latex[length=2mm]}, thick},
  pathedge/.style={-{Latex[length=2mm]}, very thick, red!75!black}
]
\fill[gray!45] (-2.7,2.25) rectangle (2.7,3.15);
\fill[gray!6]  (-2.7,1.35) rectangle (2.7,2.25);
\fill[gray!45] (-2.7,0.45) rectangle (2.7,1.35);
\fill[gray!6]  (-2.7,-0.45) rectangle (2.7,0.45);
\foreach \y in {-0.45,0.45,1.35,2.25,3.15} {\draw[black!75, line width=0.7pt] (-2.7,\y) -- (2.7,\y);}

\node[anchor=east, font=\footnotesize] at (-2.7,2.7) {$h{=}3$};
\node[anchor=east, font=\footnotesize] at (-2.7,1.8) {$h{=}2$};
\node[anchor=east, font=\footnotesize] at (-2.7,0.9) {$h{=}1$};
\node[anchor=east, font=\footnotesize] at (-2.7,0)   {$h{=}0$};

\node[state]   (s0) at (0,2.7) {$s_0$};
\node[state]   (a)  at (-1.1,2.7) {$a$};
\node[state]   (b)  at ( 1.6,1.8) {$b$};
\node[state]   (c)  at (-2.25,2.7) {$c$};
\node[state]   (d)  at ( 1.6,0.9) {$d$};
\node[goal]    (g)  at ( 0,0) {$g$};

\draw[pathedge] (s0) -- (b);
\draw[pathedge] (b)  -- (d);
\draw[pathedge] (d)  -- (g);

\draw[arrow] (0,3.55) -- (s0);
\draw[arrow] (s0) -- (a);
\draw[arrow] (a)  -- (c);
\draw[arrow] (c)  -- (g);
\end{tikzpicture}
\caption{Direct heuristic: the same state space with a different heuristic admits a strictly improving path ($s_0\!\to\!b\!\to\!d\!\to\!g$), but state $a$ is a plateau ($h(a)=h(s_0)$).}
\label{fig:direct}
\end{subfigure}
\caption{Example descending and direct heuristics for the same state space. Nodes are placed on horizontal heuristic layers; downward transitions strictly decrease the heuristic value $h$. State $s_0$ is the initial state and $g$ is the goal state. Both heuristics are dead-end avoiding since all states are solvable. The left heuristic is descending and thus also direct; the right heuristic is direct but not descending.}
\label{fig:desc-vs-effdesc}
\end{figure}

\emph{Classical planning} asks for a sequence of deterministic actions, a \emph{plan}, that
transforms a given initial state into a state satisfying a goal in a fully-observable, discrete environment with a single agent. Planning tasks are commonly described in the Planning Domain Definition Language (PDDL) \cite{mcdermott-aimag2000,haslum-et-al-2019}, and our work directly uses this representation.

A PDDL task is defined by a set of \emph{objects} and \emph{predicates}.
Together they form \emph{ground atoms}, i.e., predicates instantiated with
concrete objects, which describe the current state.
\emph{Actions} have preconditions that must hold for the action to be
applicable, and effects that add or remove ground atoms from the state.
The task also specifies an \emph{initial state}~$s_0$ and a \emph{goal}.
PDDL separates the \emph{domain} (shared actions and predicates) from the
\emph{task} (specific objects, initial state, and goal), typically stored in
two separate files.

We consider \emph{satisficing planning}, where any plan is acceptable regardless of its length, and we assume all actions have unit cost.
Most satisficing planners are based on \emph{greedy best-first search} (GBFS)~\cite{pohl-mi1969,bonet-geffner-aij2001,hoffmann-nebel-jair2001,helmert-jair2006,richter-westphal-jair2010,lipovetzky-geffner-aaai2017}, a state-space search algorithm guided by a \emph{heuristic function} $h$ that maps each state $s$ to a value $h(s) \in \mathbb{R} \cup \{\infty\}$ estimating the cost to reach a goal from $s$~\cite{pearl-1984}. Infinite values indicate that the goal is unreachable from $s$. GBFS starts from $s_0$ and maintains a priority queue of generated states ordered by their heuristic value. At each step, it \emph{expands} the most promising state $s$, i.e., the one with the smallest heuristic value $h(s)$, generating its successors $\suc(s)$ and adding them to the open list. This process continues until a goal state is reached. GBFS is complete: if a solution exists, it will eventually find one, regardless of the heuristic, as long as all generated states are eventually expanded

In contrast to the global nature of GBFS, \emph{hill climbing} (HC) is a simple local-search algorithm: starting from the initial state, it greedily moves to a successor with a strictly smaller heuristic value and terminates when it reaches a goal state. Because HC keeps no open list, it is very memory efficient. However, HC is incomplete: it can fail at \emph{plateaus} (where no successor strictly improves the heuristic) or \emph{local minima} (where every successor has a larger heuristic value), even when the goal is reachable. Completeness of HC therefore depends on the heuristic.

A \emph{descending and dead-end avoiding} heuristic~\cite{seipp-et-al-ijcai2016} eliminates both failure modes by construction. Following \citet{seipp-et-al-ijcai2016}, we call a state \emph{alive} if it is reachable, solvable, and not a goal.

\begin{definition}[Descending and Dead-End Avoiding Heuristics \cite{seipp-et-al-ijcai2016,helmert-et-al-icaps2022}]
A heuristic $h$ is \emph{descending} for a planning task $\Pi$ if every alive state $s$ has at least one successor $s' \in \suc(s)$ with $h(s') < h(s)$, and it is \emph{dead-end avoiding} if every such improving successor is solvable.
\end{definition}

\begin{proposition}[\citet{seipp-et-al-ijcai2016}]
Let $h$ be a descending, dead-end avoiding heuristic for a solvable planning task $\Pi$.
Then hill climbing guided by $h$ reaches the goal from $s_0$.
\end{proposition}

The proof is straightforward: every reachable non-goal state has an improving successor, so heuristic values strictly decrease at each step. Strict decrease prevents revisiting a state, and since the reachable state space is finite, the search must reach a goal in finitely many steps.

The DDA property is, however, unnecessarily strong. It enforces that all alive
states have improving successors, \emph{even when these states would never be
expanded} by a planner. For example, the heuristic $h$ in Figure~\ref{fig:desc-descending} is DDA because all alive states have improving
successors. But this includes $a$, which is never expanded by GBFS or
HC using $h$, because $h(b) < h(a)$, and $b$ already has a strictly improving path to the goal (i.e., $h$ strictly decreases at every step).
To address this issue, \citet{dold-helmert-aaai2024} introduce a relaxation that only considers states
reachable from $s_0$ by repeatedly taking strictly improving
transitions. They call this property \emph{wet descending and dead-end avoiding} (WDDA). For brevity, we refer to a heuristic with this property as a \emph{direct} heuristic throughout the rest of the paper.

\begin{definition}[Direct Heuristic~\cite{dold-helmert-aaai2024}]
Let $S_\downarrow$ be the set of alive states reachable from $s_0$ via strictly improving transitions. A heuristic $h$ is \emph{direct} if every $s \in S_\downarrow$ has at least one successor $s' \in \suc(s)$ with $h(s') < h(s)$, and every such improving successor is either a goal state or itself in $S_\downarrow$.
\end{definition}
For instance, state $a$ in Figure~\ref{fig:direct} is alive but has $h(a) = h(s_0)$, so $a \notin S_\downarrow$ and the direct property places no requirement on it, whereas DDA would still demand an improving successor. This seemingly minor relaxation can make the property substantially easier to satisfy and to synthesize~\cite{dold-helmert-aaai2024}.
DDA implies directness, but not the other way round.
HC guided by a direct heuristic also reaches a goal state without ever getting stuck.

Synthesizing a direct heuristic for a single PDDL task is also theoretically hard. A direct heuristic exists for a task $\Pi$ if and only if $\Pi$ is solvable: any solvable task has a direct heuristic (e.g., the perfect heuristic), and any direct heuristic guides hill climbing to a goal. Deciding whether a direct heuristic exists is therefore equivalent to deciding plan existence, which is \expspace-hard for PDDL planning (a.k.a., ``lifted planning'')~\cite{erol-et-al-aij1995,correa-degiacomo-ijcai2024}. We sidestep this per-task cost by synthesizing a single \emph{domain-general} direct heuristic from a small training set, amortizing the synthesis effort over the potentially infinite distribution of tasks in the domain.

\section{Property-Guided LLM Heuristic Function Synthesis}
\label{sec:approach}

We use an LLM to synthesize a domain-dependent heuristic function, implemented
as Python code, for a given PDDL domain. Our approach uses a single counterexample-driven repair loop: the LLM generates one candidate, we validate it on a set of training tasks, and if it fails we return a concrete counterexample to the LLM and ask for a new candidate. This loop repeats until the heuristic passes the full training set or a generation budget is exhausted. The following subsections discuss the property we target (Section~\ref{sec:eff-desc}), the prompts used in the repair loop (Section~\ref{sec:repair-loop}), and the validation procedure (Section~\ref{sec:validation}).

\subsection{Targeting Direct Heuristics}
\label{sec:eff-desc}

The DDA property is a strict global requirement: it must hold at every alive
state, and a single violating state is enough to fail, even at states that
hill climbing would never reach. We therefore target heuristics with the \emph{direct} property:
every alive state reachable from $s_0$ by strictly improving transitions must have a
strictly improving successor. A direct heuristic still guarantees that hill
climbing reaches a goal, but it focuses only on the subgraph
that hill climbing can actually traverse.

Crucially for our method, every direct-property failure yields a concrete,
local counterexample. The first type of failure occurs at a state $s$ on some strictly improving path from $s_0$ whose successors $s'$ all satisfy $h(s') \ge h(s)$; the counterexample reports $s$, its heuristic value $h(s)$, and the list of successors together with their heuristic values.
In domains with dead ends, a second type of failure can occur when the DFS reaches a non-goal state $s$ with no applicable action via a strictly improving transition from a parent $s'$; the counterexample reports $s$, its heuristic value $h(s)$, and the parent's heuristic value $h(s')$, together with the actionable suggestion that $s$ should be assigned a value $\ge h(s')$ so that hill climbing never enters this dead end.
This information is directly actionable and gives the LLM precise feedback on
how to improve the heuristic. Although validation is confined to
the training tasks, we aim for the synthesized heuristic to generalize and
remain direct on the much harder, out-of-distribution test tasks.

\subsection{Counterexample-Driven Repair Loop}
\label{sec:repair-loop}

The repair loop maintains a history of all previously generated heuristics and
their failure feedback. It starts with an \emph{initial prompt} that requests
the first candidate and uses a \emph{repair prompt} after each validation
failure.

\paragraph{Initial Prompt.}
We use the same prompt as \citet{correa-et-al-neurips2025} to make a direct
comparison easier. It provides the PDDL domain, the smallest and largest
training PDDL tasks from this domain, heuristic Python code examples for two domains disjoint from the evaluated domains, relevant planner code, and a checklist. Note that the prompt does not instruct the LLM to generate a direct heuristic.

\paragraph{Repair Prompt.}
When a candidate fails validation on a training task, the repair prompt asks
the LLM to propose a heuristic that is direct on this task as well, and
provides the definition of the direct property. It also includes:
\begin{enumerate} \denselist
    \item the PDDL domain file;
    \item the PDDL task on which validation failed;
    \item the failure feedback: the failing state, its heuristic value, and
          the successor states with their heuristic values;
    \item all previously generated heuristic functions, together with their
          failure feedback;
    \item a checklist similar to the one in the initial prompt.
\end{enumerate}

Items 2 and 3 give the LLM the exact nature of the failure.
Item 4 lets the LLM reason about what has already been tried and avoid
repeating the same mistakes (see Appendix~\ref{app:prompt} for an example prompt).

\subsection{Validation}
\label{sec:validation}

To check whether a candidate heuristic $h$ satisfies the direct property on
a task $\Pi$, we run a depth-first search (DFS) from $s_0$. At each
expanded non-goal state $s$, we check whether it has at least one successor
$s'$ with $h(s') < h(s)$. If not, the direct property fails; we record $s$, its heuristic value, and all
its successor states with their heuristic values as the counterexample. Otherwise, the DFS pushes onto the stack only the successors
$s'$ with $h(s') < h(s)$.

We apply this check to each task in the training set. As soon as a
failure is found, we return the counterexample and stop, without checking the
remaining tasks. We consider the heuristic a success if all expanded states
have a strictly improving successor across all training tasks.

Even modestly sized planning tasks can admit exponentially many strictly improving paths to a goal, so validation cannot be expected to finish within a reasonable time on every task.
We therefore impose a per-task time limit during validation.
If this limit is reached before a counterexample is found, we treat the heuristic as direct on the task and continue with the next task.

\section{Experimental Results}
\label{sec:hill-climbing}

\begin{figure}[t]
\centering
\pgfplotsset{
  trainstyle/.style={blue, very thick, mark=*, mark options={fill=blue, scale=1.1}},
  teststyle/.style={red!80!black, very thick, mark=*, mark options={fill=red!80!black, scale=1.1}},
  domainaxis/.style={
    width=1.0\textwidth,
    height=4.5cm,
    scale only axis=false,
    xmin=-30, xmax=1120,
    xtick={0,200,400,600,800,1000},
    ytick={1,2,3,4,5},
    ymin=0.5, ymax=5.5,
    xmajorgrids=true,
    ymajorgrids=true,
    grid style={dashed, gray!30},
    tick align=outside,
    tick pos=left,
    axis line style={-},
  },
}%
{\small
  \tikz[baseline=0.5ex]{%
    \draw[blue, very thick](0,0.5ex)--(1.5em,0.5ex);
    \fill[blue](0.75em,0.5ex)circle(2.2pt);}~Training
  \qquad
  \tikz[baseline=0.5ex]{%
    \draw[red!80!black, very thick](0,0.5ex)--(1.5em,0.5ex);
    \fill[red!80!black](0.75em,0.5ex)circle(2.2pt);}~Testing
}\\[4pt]
\begin{subfigure}[t]{0.485\linewidth}
\centering
\begin{tikzpicture}
\begin{axis}[
    domainaxis,
    yticklabels={Miconic, Floortile, Ferry, Childsnack, Blocks.},
]

\addplot+[trainstyle, forget plot] coordinates {(2,5.18) (29,5.18)};
\node[anchor=east, font=\scriptsize, blue]         at (axis cs:2,  5.18) {2\,};
\node[anchor=west, font=\scriptsize, blue]         at (axis cs:29, 5.18) {\,29};
\addplot+[teststyle, forget plot]  coordinates {(5,4.82) (488,4.82)};
\node[anchor=east, font=\scriptsize, red!80!black] at (axis cs:5,   4.82) {5\,};
\node[anchor=west, font=\scriptsize, red!80!black] at (axis cs:488, 4.82) {\,488};

\addplot+[trainstyle, forget plot] coordinates {(1,4.18) (10,4.18)};
\node[anchor=east, font=\scriptsize, blue]         at (axis cs:1,  4.18) {1\,};
\node[anchor=west, font=\scriptsize, blue]         at (axis cs:10, 4.18) {\,10};
\addplot+[teststyle, forget plot]  coordinates {(4,3.82) (292,3.82)};
\node[anchor=east, font=\scriptsize, red!80!black] at (axis cs:4,   3.82) {4\,};
\node[anchor=west, font=\scriptsize, red!80!black] at (axis cs:292, 3.82) {\,292};

\addplot+[trainstyle, forget plot] coordinates {(1,3.18) (20,3.18)};
\node[anchor=east, font=\scriptsize, blue]         at (axis cs:1,  3.18) {1\,};
\node[anchor=west, font=\scriptsize, blue]         at (axis cs:20, 3.18) {\,20};
\addplot+[teststyle, forget plot]  coordinates {(2,2.82) (974,2.82)};
\node[anchor=east, font=\scriptsize, red!80!black] at (axis cs:2,   2.82) {2\,};
\node[anchor=west, font=\scriptsize, red!80!black] at (axis cs:974, 2.82) {\,974};

\addplot+[trainstyle, forget plot] coordinates {(2,2.18) (30,2.18)};
\node[anchor=east, font=\scriptsize, blue]         at (axis cs:2,  2.18) {2\,};
\node[anchor=west, font=\scriptsize, blue]         at (axis cs:30, 2.18) {\,30};
\addplot+[teststyle, forget plot]  coordinates {(12,1.82) (980,1.82)};
\node[anchor=east, font=\scriptsize, red!80!black] at (axis cs:12,  1.82) {12\,};
\node[anchor=west, font=\scriptsize, red!80!black] at (axis cs:980, 1.82) {\,980};

\addplot+[trainstyle, forget plot] coordinates {(1,1.18) (10,1.18)};
\node[anchor=east, font=\scriptsize, blue]         at (axis cs:1,  1.18) {1\,};
\node[anchor=west, font=\scriptsize, blue]         at (axis cs:10, 1.18) {\,10};
\addplot+[teststyle, forget plot]  coordinates {(1,0.82) (485,0.82)};
\node[anchor=east, font=\scriptsize, red!80!black] at (axis cs:1,   0.82) {1\,};
\node[anchor=west, font=\scriptsize, red!80!black] at (axis cs:485, 0.82) {\,485};

\end{axis}
\end{tikzpicture}
\end{subfigure}%
\hfill%
\begin{subfigure}[t]{0.485\linewidth}
\centering
\begin{tikzpicture}
\begin{axis}[
    domainaxis,
    yticklabels={Transp., Spanner, Sokoban, Satellite, Rovers},
]

\addplot+[trainstyle, forget plot] coordinates {(1,5.18) (4,5.18)};
\node[anchor=east, font=\scriptsize, blue]         at (axis cs:1, 5.18) {1\,};
\node[anchor=west, font=\scriptsize, blue]         at (axis cs:4, 5.18) {\,4};
\addplot+[teststyle, forget plot]  coordinates {(1,4.82) (30,4.82)};
\node[anchor=east, font=\scriptsize, red!80!black] at (axis cs:1,  4.82) {1\,};
\node[anchor=west, font=\scriptsize, red!80!black] at (axis cs:30, 4.82) {\,30};

\addplot+[trainstyle, forget plot] coordinates {(1,4.18) (10,4.18)};
\node[anchor=east, font=\scriptsize, blue]         at (axis cs:1,  4.18) {1\,};
\node[anchor=west, font=\scriptsize, blue]         at (axis cs:10, 4.18) {\,10};
\addplot+[teststyle, forget plot]  coordinates {(2,3.82) (99,3.82)};
\node[anchor=east, font=\scriptsize, red!80!black] at (axis cs:2,  3.82) {2\,};
\node[anchor=west, font=\scriptsize, red!80!black] at (axis cs:99, 3.82) {\,99};

\addplot+[trainstyle, forget plot] coordinates {(1,3.18) (4,3.18)};
\node[anchor=east, font=\scriptsize, blue]         at (axis cs:1, 3.18) {1\,};
\node[anchor=west, font=\scriptsize, blue]         at (axis cs:4, 3.18) {\,4};
\addplot+[teststyle, forget plot]  coordinates {(1,2.82) (79,2.82)};
\node[anchor=east, font=\scriptsize, red!80!black] at (axis cs:1,  2.82) {1\,};
\node[anchor=west, font=\scriptsize, red!80!black] at (axis cs:79, 2.82) {\,79};

\addplot+[trainstyle, forget plot] coordinates {(1,2.18) (10,2.18)};
\node[anchor=east, font=\scriptsize, blue]         at (axis cs:1,  2.18) {1\,};
\node[anchor=west, font=\scriptsize, blue]         at (axis cs:10, 2.18) {\,10};
\addplot+[teststyle, forget plot]  coordinates {(1,1.82) (487,1.82)};
\node[anchor=east, font=\scriptsize, red!80!black] at (axis cs:1,   1.82) {1\,};
\node[anchor=west, font=\scriptsize, red!80!black] at (axis cs:487, 1.82) {\,487};

\addplot+[trainstyle, forget plot] coordinates {(1,1.18) (7,1.18)};
\node[anchor=east, font=\scriptsize, blue]         at (axis cs:1, 1.18) {1\,};
\node[anchor=west, font=\scriptsize, blue]         at (axis cs:7, 1.18) {\,7};
\addplot+[teststyle, forget plot]  coordinates {(3,0.82) (50,0.82)};
\node[anchor=east, font=\scriptsize, red!80!black] at (axis cs:3,  0.82) {3\,};
\node[anchor=west, font=\scriptsize, red!80!black] at (axis cs:50, 0.82) {\,50};

\end{axis}
\end{tikzpicture}
\end{subfigure}
\par\smallskip
{\small Main domain parameter}
\caption{Size of training and test tasks measured by the main domain
  parameter. Each line spans the range of values for the
  corresponding set; maximum endpoints are annotated with their numeric values.
  The parameter counts blocks (Blocksworld), children (Childsnack), cars
  (Ferry), tiles (Floortile), passengers (Miconic), rovers (Rovers),
  satellites (Satellite), boxes (Sokoban), spanners (Spanner), and vehicles
  (Transport). The test set is out-of-distribution: in most domains the
  testing range extends well beyond the training range, and test tasks may
  also differ structurally from training tasks.}
\label{fig:domain-params}
\end{figure}
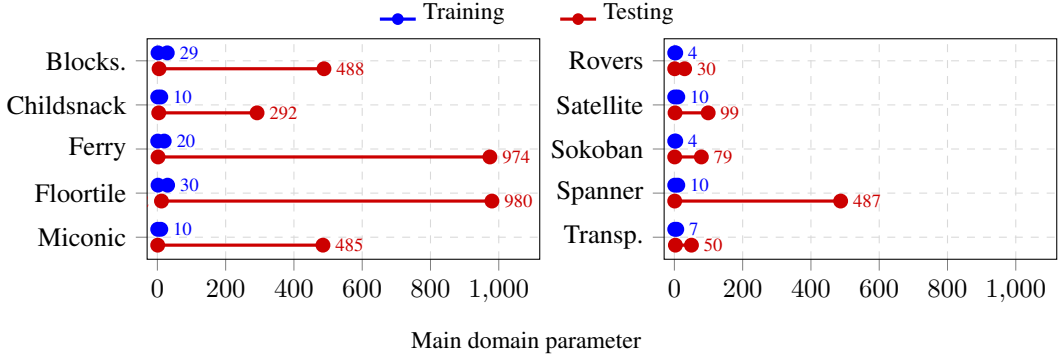

\paragraph{Experimental Setting.}
We run all experiments using Downward Lab~\cite{seipp-et-al-zenodo2017} on AMD
EPYC 7742 processors running at 2.25\,GHz. We use
Pyperplan~\cite{alkhazraji-et-al-zenodo2020} with PyPy~7.3.9 as our planning
framework for both training and testing. We generate all heuristics with
Gemini 3.1 Pro~\cite{gemini-modelcard2026}, using default API parameters. We use the ten domains together with their training and test tasks from the IPC 2023
Learning Track~\cite{taitler-et-al-aimag2024}; we summarise each domain in
Appendix~\ref{app:domains}.\footnote{Two domains, Ferry and
Satellite, use negative preconditions not supported by Pyperplan. We convert
them to equivalent domains via a transformation that eliminates
negative preconditions.} The test set contains 90 tasks per domain (900 in total)
and is out-of-distribution with respect to the training set: test tasks are
generally much larger and may differ structurally from training tasks.
Figure~\ref{fig:domain-params} shows the range of the main domain parameter
on training and testing tasks.
During training, each validation run is limited to 30\,seconds and 8\,GiB per
task, and the repair loop runs for at most 10 iterations, i.e., 10 LLM calls. During testing, each
run is limited to 5\,minutes and 8\,GiB. For our methods, we report averages over three runs.
The source code, prompts, generated heuristics, and experimental data are included as supplementary material; we will additionally release them publicly upon acceptance.

\paragraph{Scope.}
We evaluate whether property-guided synthesis reduces inference and evaluation cost while improving heuristic quality; a single representative LLM suffices to answer this question.
For a comparison of different LLMs on heuristic generation for planning, see \citet{correa-et-al-neurips2025}.
We study the LLM as a \emph{program synthesizer}: the LLM is called once per domain to generate a heuristic, which is then used to solve all tasks in that domain, including unseen test tasks. In \emph{end-to-end} planning, by contrast, the LLM is invoked separately for each training and test task, incurring much higher costs and losing all soundness guarantees on the resulting plans.
For end-to-end planning performance across LLM families, sizes, and reasoning models, see \citet{correa-et-al-neurips2025}; for frontier LLMs in this setting, see \citet{correa-e2e-arXiv2025}.
For analyses regarding data contamination, obfuscated tasks and evaluation on domains absent from LLM training data, see \citet{correa-et-al-neurips2025} and \citet{chen-arXiv2025}.

\paragraph{Baselines.} The \emph{FF} heuristic~\cite{hoffmann-nebel-jair2001} is one of the most commonly used heuristics for satisficing planning
\egcite{buechner-et-al-ipc2023b,correa-et-al-ipc2023c,gnad-et-al-ipc2023b}. It is a delete-relaxation heuristic computed in polynomial time by solving a relaxed planning problem that ignores \emph{delete effects}; it is not designed to be \emph{direct}. \inlinecite{correa-et-al-neurips2025} introduce the current state-of-the-art method for LLM-based heuristic function generation for planning:
it uses an LLM to independently generate a fixed budget of 25 candidate heuristics per domain and selects the one with the highest GBFS coverage on the training set and runs it with GBFS on the test set; it is also not designed to be direct. We refer to this method as \emph{sample-and-select} throughout (abbreviated {S\&S} in tables). We re-run this method with Gemini 3.1 Pro (the same model used by our approach) to ensure a fair comparison.

\paragraph{Inference and Evaluation Costs of Candidate Heuristics.}
Our repair loop is much more efficient than fixed-budget sample-and-select regarding the number of candidates the LLM must produce and the time spent
evaluating them.
Across the ten IPC domains and three independent runs per domain ($30$ runs
in total), our method generates $101$ candidate heuristics, on average $3.4$
per run with standard deviation $3.0$. Eight of the ten domains converge (i.e., generate a heuristic that is direct on the full training set)
within at most $5$ candidates per run, and four converge within $2$
candidates on average. Sokoban and Floortile are the clear outliers:
Sokoban averages $9$ candidates per run with a maximum of $11$ (initial
attempt plus 10 repairs), and Floortile averages $8$ with a maximum of $9$. By comparison, the sample-and-select baseline of
\citet{correa-et-al-neurips2025} fixes a budget of $25$ candidates per
domain regardless of difficulty, so the repair loop yields roughly a
$7.4\times$ reduction in inference cost per run.
Table~\ref{tab:synthesis-stats} in Appendix~\ref{app:synthesis-stats}
reports the per-domain breakdown.

Evaluation cost shrinks even more dramatically. In
sample-and-select~\cite{correa-et-al-neurips2025}, each domain evaluates a
fixed pool of candidates with GBFS, so evaluation cost
scales with the number of candidates. In our repair loop, we validate one
candidate at a time and stop at the first counterexample, which makes failed
candidates cheap to reject. The average evaluation time per domain is $10.75$\,minutes, while \citet{correa-et-al-neurips2025} report $206.25$\,CPU-hours per domain. This corresponds to roughly a $1.15\times 10^3$ reduction in evaluation cost.

\paragraph{Direct Heuristics vs.\ State of the Art.}

\begin{table}[tb]
\caption{Per-domain coverage (number of solved tasks) on the 900 IPC 2023 test tasks.
  \textbf{FF} is the classical delete-relaxation heuristic; \textbf{S\&S} is our re-run of the sample-and-select baseline~\cite{correa-et-al-neurips2025} with Gemini 3.1 Pro.
  Both baselines are run using GBFS and HC.
  \textbf{Direct} is our proposed method (direct heuristics run with HC).
  \textbf{Cov.}\ is the ablation that replaces the direct-property check with numeric coverage feedback (see ``Ablation'' paragraph), run using GBFS and HC.
  All proposed and ablation values are averages over three independent runs.
  Bold marks the best value per domain (ties are co-bolded).
  The full table with standard deviations is in Appendix~\ref{app:full-coverage} (Table~\ref{tab:coverage-by-domain-full}).}
\label{tab:coverage-by-domain}
\centering
\setlength{\tabcolsep}{5pt}
\begin{tabular}{@{}lrrrrrrr@{}}
\toprule
\multicolumn{1}{c}{} & \multicolumn{4}{c}{\textbf{Baselines}} & \multicolumn{2}{c}{\textbf{Ablation}} & \multicolumn{1}{c}{\textbf{Ours}} \\
\cmidrule(lr){2-5}\cmidrule(lr){6-7}\cmidrule(lr){8-8}
\multicolumn{1}{c}{} & \multicolumn{2}{c}{\textbf{FF}} & \multicolumn{2}{c}{\textbf{S\&S}} & \multicolumn{2}{c}{\textbf{Cov.}} & \multicolumn{1}{c}{\textbf{~~Direct}} \\
\cmidrule(lr){2-3}\cmidrule(lr){4-5}\cmidrule(lr){6-7}\cmidrule(l){8-8}
\textbf{Domain} & \textbf{GBFS} & \textbf{HC} & \textbf{GBFS} & \textbf{HC} & \textbf{GBFS} & \textbf{HC} & \textbf{HC} \\
\midrule
Blocksworld (90)    & 23  & 0  & 84          & \textbf{90} & 81.0  & 49.7  & 77.3          \\
Childsnack (90)     & 12  & 0  & 58          & 70          & 58.0  & 46.7  & \textbf{71.7} \\
Ferry (90)          & 45  & 0  & 67          & 71          & 64.7  & 49.3  & \textbf{72.3} \\
Floortile (90)      & 11  & 0  & 30          & 0           & 39.0  & 15.7  & \textbf{40.7} \\
Miconic (90)        & 61  & 1  & 83          & 7           & 82.0  & 34.0  & \textbf{90.0} \\
Rovers (90)         & 29  & 2  & \textbf{44} & 12          & 36.7  & 20.7  & 42.7          \\
Satellite (90)      & 35  & 17 & 57          & \textbf{60} & 56.0  & 57.3  & 55.7 \\
Sokoban (90)        & 29  & 1  & \textbf{30} & 0           & 26.0  & 12.7  & \textbf{30.0} \\
Spanner (90)        & 30  & 6  & 66          & \textbf{79} & 66.0  & \textbf{79.0} & 75.0  \\
Transport (90)      & 23  & 3  & 54          & 41          & 62.7  & 47.3  & \textbf{68.0} \\
\midrule
\textbf{Total (900)} & 298 & 30 & 573         & 430         & 572.7 & 412.3 & \textbf{623.3} \\
\bottomrule
\end{tabular}
\end{table}

\begin{figure}[tb]
\centering
\begin{subfigure}[t]{0.366\linewidth}
\centering
\includegraphics[width=\linewidth]{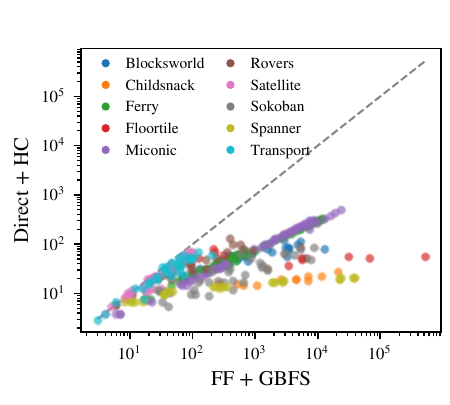}
\phantomcaption\label{fig:effdesc-vs-ff}
\end{subfigure}\hfill
\begin{subfigure}[t]{0.317\linewidth}
\includegraphics[width=\linewidth]{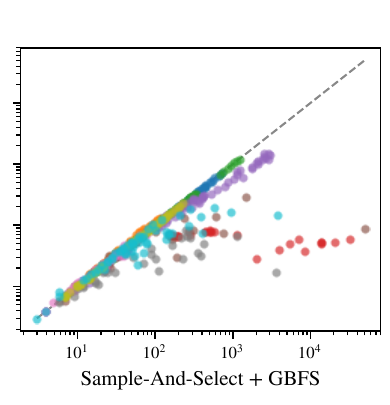}
\phantomcaption\label{fig:effdesc-vs-sampling}
\end{subfigure}\hfill
\begin{subfigure}[t]{0.317\linewidth}
\centering
\includegraphics[width=\linewidth]{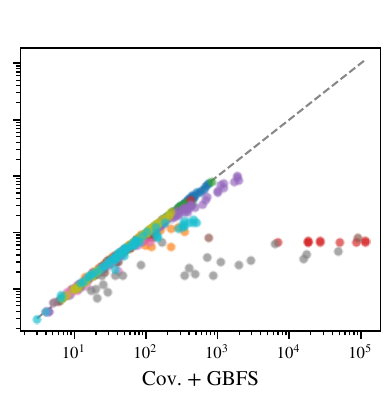}
\phantomcaption\label{fig:effdesc-vs-coverage}
\end{subfigure}
\vspace{-15pt}
\caption{State expansions of direct + HC versus each baseline, on test tasks solved by both methods. Each point is one task, colored by domain. Points below the diagonal favor direct + HC. Note that restricting to tasks solved by both methods, in general, favors the method that solves fewer tasks.}
\label{fig:effdesc-expansions}
\end{figure}

We compare the direct heuristics to the two baselines: the FF
heuristic~\cite{hoffmann-nebel-jair2001} and the sample-and-select method by
\citet{correa-et-al-neurips2025}. Both baselines use GBFS, which may
backtrack extensively before finding a solution. Our method uses HC, which
follows a single greedy path and reaches the goal without any search.
Table~\ref{tab:coverage-by-domain} reports results for the baselines under both GBFS and HC.

The main result is that our direct heuristics with HC solve
$623.3$ tasks by going directly to the goal, while sample-and-select with
GBFS solves $573$ tasks, and FF with GBFS
solves only $298$.
That is, we solve more tasks than the best prior
LLM method, and we do so without any search. Running the sample-and-select
heuristics under HC solves only $430$ tasks, confirming that the gain comes
from the property our repair loop trains for, not merely from
the choice of search algorithm. FF under HC collapses to $30$ solved tasks. It is not
a descending heuristic and offers no guarantee of progress at every state, so
HC fails at the first plateau or local minimum it encounters.
Figures~\ref{fig:effdesc-vs-ff} and \ref{fig:effdesc-vs-sampling} make the efficiency gap concrete: on tasks solved by both methods, our heuristics with HC expand far fewer states than sample-and-select, sometimes by several orders of magnitude. Together with the $7.4\times$ reduction in inference cost and the $1.15\times 10^3$ reduction in evaluation cost reported above, these results show that property-guided synthesis is generally better than sample-and-select: it solves more tasks, reaches the goal with fewer expansions, and costs far less to synthesize.

\paragraph{Generalization of the Direct Property.}
We validate the direct property only on small training tasks, and there is no guarantee that a heuristic that passes this check will remain direct on the much larger, out-of-distribution test tasks. At test time, hill climbing fails as soon as it reaches a state~$s$ with no successor~$s'$ satisfying $h(s') < h(s)$, which is exactly a violation of the direct property. Across the three independent runs over all 900 test tasks ($2{,}700$ evaluations in total), only a single evaluation stops for this reason (in Floortile). The remaining evaluations either find a plan or hit the time limit. The synthesized heuristics are therefore effectively direct on the out-of-distribution test set as well. This is especially striking in Sokoban, a \pspace-complete domain~\cite{culberson-tr1997}, where no test evaluation across any of the three runs ever fails because of a property violation. The heuristics we generate from small training tasks therefore generalize robustly to much harder test tasks.

\paragraph{Ablation.}
To isolate the contribution of the direct property, we run the same repair loop but replace the property check with a numeric coverage score: the repair prompt reports which training task the heuristic failed to solve within the time limit and how many states were expanded before timing out, and asks for a new heuristic that solves more tasks with GBFS. We refer to this ablation as \emph{Cov.}\ Under GBFS, Cov.\ solves $572.7$ ($\pm 15.95$) tasks, roughly matching sample-and-select ($573$). Under HC, however, Cov.\ degrades sharply to $412.3$ tasks, well below our direct heuristics with HC ($623.3$, $\pm 6.7$). Coverage feedback alone is therefore not sufficient to obtain heuristics that guide HC. Figure~\ref{fig:effdesc-vs-coverage} reinforces this point: on tasks solved by both methods, direct heuristics consistently require fewer expansions.

\paragraph{Qualitative Analysis.}
\label{sec:heuristic-cases}
Inspecting what the repair loop changes in the heuristic reveals why the direct property is a useful synthesis target. Two domains illustrate the outcomes.

\emph{Blocksworld.}
Blocksworld has a single arm that rearranges blocks into goal stacks by picking up, putting down, stacking, and unstacking one block at a time.
The initial heuristic counts 2 actions per misplaced block (one to pick it up, one to put it down), subtracts 1 if a block is currently held, and adds 2 for each deadlock it detects.
It uses two different deadlock conditions: for blocks on the table, the penalty activates only when the target block is itself misplaced; for a held block, the penalty activates whenever its target is not ready, meaning the target is either misplaced or covered by another block.
Consider the state where $b_1$ is on $b_2$ and $b_3$ is on $b_4$, with the goal of swapping them: $b_1$ should go on $b_4$ and $b_3$ on $b_2$, while $b_2$ and $b_4$ are already correctly on the table.
Both $b_1$ and $b_3$ are misplaced, giving $h = 4$.
Since $b_2$ and $b_4$ are correctly on the table, the first condition does not activate and no penalty is added.
After unstacking $b_3$, the arm holds $b_3$ and needs to place it on $b_2$, but $b_1$ still covers $b_2$, so $b_2$ is not ready: the second condition activates and adds $+2$, giving $h = 2 \times 2 - 1 + 2 = 5$.
The same happens when unstacking $b_1$.
The inconsistency is that the arm-empty state and the holding state use different conditions, so the mutual blocking between $b_1$ and $b_3$ is invisible in the arm-empty state but penalized in every successor, creating a local minimum.
The final heuristic fixes this by building a dependency graph among misplaced blocks.
It adds an edge from $b_3$ to $b_1$ (because $b_3$ covers $b_4$, the target of $b_1$) and from $b_1$ to $b_3$ (because $b_1$ covers $b_2$, the target of $b_3$), detects the cycle, and adds $+2$ in the arm-empty state itself, giving $h = 6$.
The successors then give $h = 5$, restoring descent.
The final heuristic is effectively direct on all test tasks.

\emph{Satellite.}
Satellite requires a satellite to switch on an instrument, turn to one of its calibration targets, calibrate it, and then point to the image direction before taking the image.
The initial heuristic estimates each missing image goal independently by adding costs for switching on an instrument, calibrating it, turning to the needed directions, and taking the image.
This gives an estimate of goal distance, but it ignores the ordering constraints created by calibration and by the single power slot on each satellite.
An action may temporarily move the satellite away from the image goal direction, and switching from one instrument to another can force recalibration.
The final heuristic fixes this by reasoning separately about each satellite's remaining work as a short action sequence.
For each satellite, it groups the images that still need to be taken, considers which instrument should achieve them, and counts the actions still needed to finish them in order: switching instruments, calibrating, turning, and taking the images.
It no longer treats the remaining requirements as an unordered set of subgoals, but as a sequence whose order matters.
Because of this, each improving move removes a real remaining step in that sequence, which is why the final heuristic becomes direct.

\section{Related Work}

\paragraph{Counterexample-Guided Inductive Synthesis.}
Counterexample-guided inductive synthesis (CEGIS)~\cite{solar-lezama-pldi2005,jha-icse2010,abate-cav2018} alternates between a synthesizer that proposes candidate programs and a verifier that checks them against a formal specification.
When a candidate fails, the verifier returns a concrete counterexample, and the synthesizer uses it to rule out or repair the candidate.
Our work and that of \citet{orvalho-aaai2025} are recent examples of CEGIS-style counterexample feedback applied to LLM-based program synthesis.

\paragraph{LLM-Based Program Synthesis.}
Most LLM-based program synthesis evaluates candidates with numeric scores such as test-pass rates or the value of the produced solutions, and uses these scores to rank or select among many candidates~\cite{li-et-al-arxiv2022,liu-icml2024,tuisov-et-al-icaps2026,chen-arXiv2025,romera-paredes-et-al-nature2024,novikov2025-arXiv2025}.
In general, these approaches can combine multiple scores and use specific prompt strategies to guide the LLM toward better candidates, but they do not provide concrete feedback on how to improve a candidate.

\paragraph{LLM-Based Heuristic Generation.}
Our closest precursor is \citet{correa-et-al-neurips2025}, who sample a fixed budget of heuristic candidates with an LLM and select the one that solves the most training tasks. Other recent work that generates heuristics with LLMs includes \citet{tuisov-et-al-icaps2026}, who focus on a different planning formalism, and \citet{chen-arXiv2025}, who focus on generalized policies.

\paragraph{Descending Heuristics.}
The descending and dead-end avoiding (DDA) property was coined by \citet{seipp-et-al-ijcai2016}.
\citet{helmert-et-al-icaps2022} showed that verifying or synthesizing DDA potential heuristics~\cite{pommerening-et-al-aaai2015} is \pspace-complete for planning tasks encoded in propositional logic, with structural relaxations dropping the complexity to lower levels of the polynomial hierarchy.
\citet{dold-helmert-aaai2024} introduce the direct (orig. ``wet DDA'') variant, which only requires descent along states reachable from $s_0$ by strictly improving transitions. These properties are useful to estimate the hardness of planning tasks based on the simplest form of their DDA/direct heuristics \cite{seipp-et-al-ijcai2016,dold-helmert-aaai2024}.

\paragraph{Learning Search Control.}
\citet{frances-et-al-ijcai2019} synthesize generalized potential heuristics as weighted sums over description-logic features \cite{baader-et-al-2017} and prove DDA on the covered domains; their approach is restricted to domains admitting a DDA heuristic representable in their specific feature class and is therefore not directly comparable to ours.
Several lines of work synthesize or learn search control mechanisms for planning domains~\cite{bonet-et-al-aaai2019,drexler-et-al-icaps2022,drexler-et-al-jair2024}. There is also a long history of approaches that learn heuristics for planning~\cite{samuel-ibm1959,christensen-et-al-aaai1986,samadi-et-al-aaai2008,arfaee-et-al-aij2011}. These include methods that learn for specific tasks~\cite{ferber-et-al-icaps2022,oToole-et-al-socs2022,bettker-et-al-jair2024} and methods that learn for entire domains~\cite{shen-et-al-icaps2020,stahlberg-et-al-icaps2022,chen-et-al-aaai2024,hao-et-al-ijcai2024}.

\paragraph{LLMs for Planning.}
In addition, there is a large body of recent work that uses LLMs for planning in several settings such as end-to-end plan generation~\cite{bohnet-et-al-arxiv2024,stechly-et-al-neurips2024,huang-et-al-arxiv2024}, PDDL domain or task construction~\cite{tantakoun-etal-acl2025,guan-et-al-neurips2023,oswald-et-al-icaps2024,bo-et-al-arxiv2024}, and generalized planning~\cite{rossetti-et-al-icaps2024,murray-et-al-icaps2026}.

\section{Limitations}

Our method depends on the existence of a property that can be efficiently checked and turned into actionable feedback. In planning, the direct property provides exactly this. For problems where no such informative property is available, or where the training set is too small to expose meaningful counterexamples, the repair loop loses its main advantage and degenerates to score-based ranking.

A second limitation concerns the scope of verification. We check the direct property only on a small training set under bounded resources, so passing this check does not certify that the heuristic is direct on unseen tasks. Our experiments show that the property typically generalizes well to much larger out-of-distribution tasks, but counterexamples on test tasks remain possible in principle.

A third limitation is that we synthesize unconstrained Python code. This flexibility is useful, but it makes the resulting heuristics harder to interpret and verify than structured representations such as fixed feature sets or linear potentials. A promising direction for future work is to combine our repair loop with strategies that automatically prove the synthesized heuristic to be direct on an entire domain, as has been done in restricted settings~\cite{abdulaziz-et-al-icaps2022wshsdip}, possibly using theorem provers powered by LLMs.

\section{Conclusions}

In this paper, we studied property-guided LLM program synthesis for planning and showed that a counterexample-driven repair loop can synthesize direct heuristic functions efficiently. Across the ten IPC 2023 Learning Track domains, and compared to the previous best heuristic generation method, our method requires about $7.4\times$ fewer LLM calls for heuristic generation, reduces candidate evaluation cost by roughly $1.15\times 10^3$, and solves more out-of-distribution test tasks by hill climbing directly to the goal. The direct property is effective both as a synthesis target and as a feedback mechanism: when violated, it provides concrete counterexamples that guide repair, and when satisfied on the training tasks, it often generalizes to much larger unseen tasks. More broadly, our results suggest that whenever a synthesis problem admits a checkable property with actionable counterexamples, property-guided LLM repair can be a practical alternative to score-based candidate ranking.

\section*{Acknowledgments}

This work was supported by the Wallenberg AI, Autonomous Systems and Software Program (WASP) funded by the Knut and Alice Wallenberg Foundation.
André G.~Pereira acknowledges support from FAPERGS with projects 21/2551-0000741-9 and 25/2551-0002590-7.
This study was financed in part by the \textit{Coordenação de Aperfeiçoamento de Pessoal de Nível Superior -- Brasil} (CAPES) -- Finance Code~001.

\bibliographystyle{plainnat}
\bibliography{abbrv,literatur,biblio,crossref-short}

\newpage
\appendix

\section{Benchmark Domains}
\label{app:domains}

We evaluate on the ten domains of the Learning Track of the IPC
2023~\cite{taitler-et-al-aimag2024}. We give a brief description of each
domain below, based on the PDDL definitions distributed by the competition.

\paragraph{Blocksworld.}
A single arm rearranges distinguishable blocks into a goal configuration of
stacks. The arm holds at most one block at a time and supports four actions:
pick a clear block up from the table, put a held block down on the table,
stack a held block onto another clear block, and unstack a clear block from
the block below it. Goals specify the desired stacking via \texttt{on} and
\texttt{on-table} atoms.

\paragraph{Childsnack.}
An agent must prepare sandwiches in a kitchen and serve them to children
seated at tables. Each child is either gluten-allergic or not; gluten-free
sandwiches require both gluten-free bread and gluten-free content, both of
which are scarce. Sandwiches are placed on trays that move between the
kitchen and the tables. The goal is to have every child served a sandwich
compatible with their allergy.

\paragraph{Ferry.}
A single ferry transports cars between locations on a sea. The ferry carries
at most one car at a time and alternates between sailing (empty or loaded),
boarding, and disembarking. The goal specifies the target location of each
car.

\paragraph{Floortile.}
Robots paint tiles on a rectangular grid in two colors. Each robot can move
onto an adjacent unpainted tile, change the color it currently holds, or
paint the tile immediately above or below itself. A robot cannot step onto a
painted tile, so the planner must reason about painting order and about
positioning robots so they do not block each other.

\paragraph{Miconic.}
A single elevator picks up passengers at their origin floors and delivers
them to their destination floors. The four actions are board, depart, drive
up, and drive down; passenger capacity is unbounded. The goal is to have
every passenger served at their destination.

\paragraph{Rovers.}
A team of rovers performs scientific operations on a network of waypoints and
communicates results to a lander. Operations include sampling soil and rock
at the current waypoint, calibrating an on-board camera against a calibration
target, taking images of objectives, and communicating data via line of sight
to the lander. Rovers differ in what they can traverse and which instruments
they carry, so plans must allocate tasks across heterogeneous agents.

\paragraph{Satellite.}
A fleet of satellites takes calibrated images of celestial directions in
particular imaging modes. Before an image can be acquired, the relevant
instrument must be switched on, calibrated against a calibration target, and
the satellite must be pointing at the target direction. Each satellite has a
single power slot, so instruments must be switched on one at a time.

\paragraph{Sokoban.}
An agent on a grid pushes boxes onto designated goal cells. The agent can move
into a clear adjacent cell or, if a box sits in the adjacent cell, push it
one cell further in the same direction, provided the destination cell is
clear. Boxes cannot be pulled, so a box pushed against a wall or into a
corner in the wrong orientation creates an unrecoverable dead end.

\paragraph{Spanner.}
A man walks along a directed chain of locations to a workshop, where he must
tighten a number of loose nuts. Spanners are scattered along the way; each
spanner can be used to tighten exactly one nut and the chain cannot be
retraversed. The man must therefore pick up enough usable spanners before
reaching the workshop.

\paragraph{Transport.}
Trucks with bounded discrete capacity move packages along a road network.
Pick-up and drop actions consume and free one unit of capacity, encoded as a
chain of size objects related by a predecessor predicate. The plan must
coordinate truck routes and remaining capacity to deliver every package to
its goal location.

\section{Full Coverage Results with Standard Deviations}
\label{app:full-coverage}

Table~\ref{tab:coverage-by-domain-full} reports the same per-domain coverage
numbers as Table~\ref{tab:coverage-by-domain} in the main text, but additionally
includes standard deviations across the three independent runs for the
\textbf{Cov.}\ ablation (GBFS and HC) and the \textbf{Direct} proposed method (HC).
The baseline columns (\textbf{FF} and \textbf{S\&S} under GBFS and HC) are
reproduced without standard deviations because they are deterministic.

\begin{table}[ht]
\caption{Per-domain coverage (number of solved tasks) on the 900 IPC 2023 test tasks.
  This is the full version of Table~\ref{tab:coverage-by-domain} in the main text,
  including standard deviations across three independent runs for the ablation and proposed method.
  \textbf{FF} is the classical delete-relaxation heuristic;
  \textbf{S\&S} is our re-run of the sample-and-select baseline~\cite{correa-et-al-neurips2025} with Gemini 3.1 Pro.
  Both baselines are run using GBFS and HC.
  \textbf{Direct} is our proposed method (direct heuristics run with HC).
  \textbf{Cov.}\ is the ablation that replaces the direct-property check with numeric coverage feedback,
  using both GBFS and HC.
  Bold marks the best value per domain (ties are co-bolded).}
\label{tab:coverage-by-domain-full}
\centering
\setlength{\tabcolsep}{3pt}
\begin{tabular}{@{}lrrrrrrr@{}}
\toprule
\multicolumn{1}{c}{} & \multicolumn{4}{c}{\textbf{Baseline}} & \multicolumn{2}{c}{\textbf{Ablation}} & {\textbf{Our Method}} \\
\cmidrule(lr){2-5}\cmidrule(lr){6-7}\cmidrule(lr){8-8}
\multicolumn{1}{c}{} & \multicolumn{2}{c}{\textbf{FF}} & \multicolumn{2}{c}{\textbf{S\&S}} & \multicolumn{2}{c}{\textbf{Cov.}} & \multicolumn{1}{c}{\textbf{Direct}} \\
\cmidrule(lr){2-3}\cmidrule(lr){4-5}\cmidrule(lr){6-7}\cmidrule(lr){8-8}
\textbf{Domain} & \textbf{GBFS} & \textbf{HC} & \textbf{GBFS} & \textbf{HC} & \textbf{GBFS} & \textbf{HC} & \textbf{HC} \\
\midrule
Blocksworld (90)    & 23  & 0  & 84          & \textbf{90} & 81.00 $\pm$ 2.00  & 49.67 $\pm$ 3.79  & 77.3 $\pm$ 11.5          \\
Childsnack (90)     & 12  & 0  & 58          & 70          & 58.00 $\pm$ 0.00  & 46.67 $\pm$ 40.41 & \textbf{71.7} $\pm$ 0.6  \\
Ferry (90)          & 45  & 0  & 67          & 71          & 64.67 $\pm$ 4.04  & 49.33 $\pm$ 22.48 & \textbf{72.3} $\pm$ 1.2  \\
Floortile (90)      & 11  & 0  & 30          & 0           & 39.00 $\pm$ 11.36 & 15.67 $\pm$ 16.01 & \textbf{40.7} $\pm$ 9.0  \\
Miconic (90)        & 61  & 1  & 83          & 7           & 82.00 $\pm$ 8.00  & 34.00 $\pm$ 48.59 & \textbf{90.0} $\pm$ 0.0  \\
Rovers (90)         & 29  & 2  & \textbf{44} & 12          & 36.67 $\pm$ 5.69  & 20.67 $\pm$ 14.15 & 42.7 $\pm$ 13.6          \\
Satellite (90)      & 35  & 17 & 57          & \textbf{60} & 56.00 $\pm$ 1.73  & 57.33 $\pm$ 5.51  & 55.7 $\pm$ 14.5          \\
Sokoban (90)        & 29  & 1  & \textbf{30} & 0           & 26.67 $\pm$ 1.15  & 12.67 $\pm$ 6.43  & \textbf{30.0} $\pm$ 0.0  \\
Spanner (90)        & 30  & 6  & 66          & \textbf{79} & 66.00 $\pm$ 0.00  & \textbf{79.00} $\pm$ 0.00  & 75.0 $\pm$ 0.0  \\
Transport (90)      & 23  & 3  & 54          & 41          & 62.67 $\pm$ 4.04  & 47.33 $\pm$ 28.54 & \textbf{68.0} $\pm$ 0.0  \\
\midrule
\textbf{Total (900)} & 298 & 30 & 573         & 430         & 572.67 $\pm$ 15.95 & 412.33 $\pm$ 41.93 & \textbf{623.3} $\pm$ 6.7 \\
\bottomrule
\end{tabular}
\end{table}

\section{Synthesis Statistics}
\label{app:synthesis-stats}

Table~\ref{tab:synthesis-stats} summarises how many candidate direct heuristics the
repair loop generates before the property check passes on the full training
set. We report per-domain statistics over the three independent runs:
the minimum, mean (with standard deviation), and maximum number of candidates
generated in a single run, and the total over the three runs.

\begin{table}[ht]
\caption{Number of candidate direct heuristics generated by the repair loop per
synthesis run, summarised across the three independent runs per domain.
$n$ is the number of runs, and the Min, Mean $\pm$ SD, and Max columns are
computed over the per-run candidate counts.}
\label{tab:synthesis-stats}
\centering
\setlength{\tabcolsep}{6pt}
\begin{tabular}{@{}lrrrr@{}}
\toprule
\textbf{Domain} & \textbf{$n$} & \textbf{Min} & \textbf{Mean $\pm$ SD} & \textbf{Max} \\
\midrule
Blocksworld & 3 & 1 & 2.33 $\pm$ 2.31  & 5  \\
Childsnack  & 3 & 1 & 2.00 $\pm$ 1.00  & 3  \\
Ferry       & 3 & 1 & 2.00 $\pm$ 1.00  & 3  \\
Floortile   & 3 & 7 & 8.33 $\pm$ 1.15  & 9  \\
Miconic     & 3 & 1 & 1.67 $\pm$ 0.58  & 2  \\
Rovers      & 3 & 2 & 2.33 $\pm$ 0.58  & 3  \\
Satellite   & 3 & 2 & 2.33 $\pm$ 0.58  & 3  \\
Sokoban     & 3 & 5 & 9.00 $\pm$ 3.46  & 11 \\
Spanner     & 3 & 1 & 1.33 $\pm$ 0.58  & 2  \\
Transport   & 3 & 2 & 2.33 $\pm$ 0.58  & 3  \\
\midrule
\textbf{Overall} & 30 & 1 & 3.37 $\pm$ 2.99 & 11 \\
\bottomrule
\end{tabular}
\end{table}

\section{Example Repair Prompt}
\label{app:prompt}

Below we show the repair prompt used in the repair loop, instantiated for the Blocksworld domain at iteration~3.
We edit the prompt to highlight its main structural components and match the nomenclature used in the paper.

\begin{lstlisting}[basicstyle=\scriptsize\ttfamily, breaklines=true]
<role>
You are a highly-skilled AI planning researcher and a proficient Python programmer creating a domain-dependent heuristic function for the PDDL domain blocksworld.
</role>

<objective>
Produce a revised heuristic where every non-goal state reachable by Depth-First Search, which only expands strictly improving successors, has at least one improving successor, regardless of the order in which operators are applied.
</objective>

<evaluation-method>
The heuristic is validated using a Depth-First Search (DFS) that enforces the direct property at every step:
- Only successors with h(s') < h(s) are added to the DFS stack.
- A property violation occurs when DFS expands a non-goal state and finds no successor with a strictly lower heuristic value.
...
</evaluation-method>

<success-criteria>
1. Alive State: reachable from the initial state, solvable, and not a goal state.
2. Improving Successor: a successor s' such that h(s') < h(s).
3. Direct: every alive state reachable from the initial state via a sequence of strictly decreasing h values has at least one improving successor. All states that DFS can reach must contain an improving successor.
4. Dead-End Avoiding: no improving successor of an alive state reachable by DFS is itself unsolvable (h = infinity).
</success-criteria>

<previous-iterations>
Iteration 0:
- Prompt Type: Init
- Solved 5 instances (Average expanded nodes: 2.4)
- Direct validated on 5 tasks (last index: 4)
- Failure Analysis:
  Failure Type: property
  Failing Task: p06.pddl
  Property Failure Kind: direct
  State: ['(clear b1)', '(clear b2)', '(holding b3)', '(on-table b1)', '(on-table b2)']
  Heuristic Value: 3
  Reason: State has h=3 but no successor has h < 3.
  Successors:
    1. action=(stack b3 b2), h=6, added=['(clear b3)', '(on b3 b2)', '(arm-empty)'], deleted=['(holding b3)', '(clear b2)']
    2. action=(stack b3 b1), h=6, added=['(clear b3)', '(on b3 b1)', '(arm-empty)'], deleted=['(holding b3)', '(clear b1)']
    3. action=(putdown b3), h=4, added=['(clear b3)', '(on-table b3)', '(arm-empty)'], deleted=['(holding b3)']
<previous-heuristic-code>
...
</previous-heuristic-code>

Iteration 1:
- Prompt Type: Property
- Main Idea: ...
- Solved 16 instances (Average expanded nodes: 7.0)
- Direct validated on 16 tasks (last index: 15)
- Failure Analysis:
  Failure Type: property
  Failing Task: p17.pddl
  Property Failure Kind: direct
  State: ['(arm-empty)', '(clear b1)', '(clear b3)', '(clear b4)', '(on b2 b5)', '(on b3 b2)', '(on-table b1)', '(on-table b4)', '(on-table b5)']
  Heuristic Value: 10
  Reason: State has h=10 but no successor has h < 10.
  Successors:
    1. action=(unstack b3 b2), h=11, added=['(holding b3)', '(clear b2)'], deleted=['(clear b3)', '(on b3 b2)', '(arm-empty)']
    2. action=(pickup b4), h=11, added=['(holding b4)'], deleted=['(on-table b4)', '(clear b4)', '(arm-empty)']
    3. action=(pickup b1), h=11, added=['(holding b1)'], deleted=['(clear b1)', '(on-table b1)', '(arm-empty)']
<previous-heuristic-code>
...
</previous-heuristic-code>

Iteration 2:
- Prompt Type: Property
- Main Idea: ...
- Solved 17 instances (Average expanded nodes: 7.8)
- Direct validated on 17 tasks (last index: 16)
- Failure Analysis:
  Failure Type: property
  Failing Task: p18.pddl
  Property Failure Kind: direct
  State: ['(arm-empty)', '(clear b1)', '(clear b3)', '(clear b5)', '(on b2 b4)', '(on b5 b2)', '(on-table b1)', '(on-table b3)', '(on-table b4)']
  Heuristic Value: 8
  Reason: State has h=8 but no successor has h < 8.
  Successors:
    1. action=(unstack b5 b2), h=9, added=['(holding b5)', '(clear b2)'], deleted=['(on b5 b2)', '(clear b5)', '(arm-empty)']
    2. action=(pickup b1), h=9, added=['(holding b1)'], deleted=['(clear b1)', '(on-table b1)', '(arm-empty)']
    3. action=(pickup b3), h=9, added=['(holding b3)'], deleted=['(clear b3)', '(on-table b3)', '(arm-empty)']
<previous-heuristic-code>
...
</previous-heuristic-code>
</previous-iterations>

<current-status>
Domain Progress: solving 99 instances of increasing difficulty.
Validation Time Limit: 30 seconds per task.
Current Result: The heuristic was direct on 73 instances, but failed on p74.pddl.
</current-status>

<failure-analysis>
Failure Type: property
Property Failure Kind: direct
State: ['(arm-empty)', '(clear b10)', '(clear b11)', '(clear b15)', '(clear b17)', '(clear b2)', '(clear b21)', '(clear b22)', '(clear b3)', '(clear b4)', '(clear b5)', '(clear b6)', '(clear b8)', '(clear b9)', '(on b1 b14)', '(on b12 b18)', '(on b14 b13)', '(on b15 b12)', '(on b18 b7)', '(on b21 b1)', '(on b6 b20)', '(on b7 b19)', '(on b9 b16)', '(on-table b10)', '(on-table b11)', '(on-table b13)', '(on-table b16)', '(on-table b17)', '(on-table b19)', '(on-table b2)', '(on-table b20)', '(on-table b22)', '(on-table b3)', '(on-table b4)', '(on-table b5)', '(on-table b8)']
Heuristic Value: 38
Reason: State has h=38 but no successor has h < 38.
Successor Analysis:
1. action=(unstack b15 b12), h=39, added=['(holding b15)', '(clear b12)'], deleted=['(clear b15)', '(on b15 b12)', '(arm-empty)']
2. action=(unstack b9 b16), h=39, added=['(holding b9)', '(clear b16)'], deleted=['(on b9 b16)', '(clear b9)', '(arm-empty)']
3. action=(pickup b11), h=39, added=['(holding b11)'], deleted=['(on-table b11)', '(clear b11)', '(arm-empty)']
4. action=(pickup b8), h=39, added=['(holding b8)'], deleted=['(arm-empty)', '(clear b8)', '(on-table b8)']
5. action=(pickup b3), h=39, added=['(holding b3)'], deleted=['(clear b3)', '(on-table b3)', '(arm-empty)']
6. action=(pickup b4), h=39, added=['(holding b4)'], deleted=['(on-table b4)', '(clear b4)', '(arm-empty)']
7. action=(pickup b17), h=39, added=['(holding b17)'], deleted=['(clear b17)', '(on-table b17)', '(arm-empty)']
8. action=(unstack b6 b20), h=39, added=['(holding b6)', '(clear b20)'], deleted=['(clear b6)', '(on b6 b20)', '(arm-empty)']
9. action=(pickup b10), h=39, added=['(holding b10)'], deleted=['(on-table b10)', '(clear b10)', '(arm-empty)']
10. action=(pickup b22), h=39, added=['(holding b22)'], deleted=['(on-table b22)', '(clear b22)', '(arm-empty)']
11. action=(pickup b5), h=39, added=['(holding b5)'], deleted=['(on-table b5)', '(clear b5)', '(arm-empty)']
12. action=(unstack b21 b1), h=39, added=['(clear b1)', '(holding b21)'], deleted=['(clear b21)', '(on b21 b1)', '(arm-empty)']
13. action=(pickup b2), h=39, added=['(holding b2)'], deleted=['(clear b2)', '(on-table b2)', '(arm-empty)']
</failure-analysis>

This is the current Python code of the heuristic function:
<heuristic-code-file>
...
</heuristic-code-file>

This is the PDDL domain file of the blocksworld domain:
<domain-file>
...
</domain-file>

This is the first PDDL instance file of the blocksworld domain where the heuristic is failing the direct property:
<task-file>
...
</task-file>

Here is a checklist to help you with your code:
<checklist>
  1) All used modules are imported.
  2) The information from static facts is extracted into suitable data structures in the constructor.
  3) Use the failure analysis and iteration history to preserve what worked and fix what failed.
  4) The name of the heuristic should be BlocksworldHeuristic.
</checklist>

<final-instruction>
  Return a new heuristic function code that satisfies the direct property in the instance in <task-file> as well.

  Output your answer in the following XML format:
  <generated-main-idea>
    A short paragraph describing only the idea of the new heuristic you are returning. Do not include any other information.
  </generated-main-idea>
  <generated-heuristic-code>
    Provide the Python code of the domain-dependent heuristic for the blocksworld.
  </generated-heuristic-code>
</final-instruction>
\end{lstlisting}


\end{document}